\documentclass[10pt,twocolumn,letterpaper]{article}

\usepackage[pagenumbers]{iccv} 

\usepackage[accsupp]{axessibility}
\usepackage{stfloats}
\usepackage{afterpage}
\usepackage{subcaption}
\usepackage{siunitx}
\usepackage{amsmath}
\usepackage{enumitem}
\usepackage{courier}
\usepackage{multirow}
\usepackage{makecell}
\usepackage{tabularx} 
\usepackage{amssymb}
\usepackage{bm}
\usepackage{float}
\usepackage{pifont} 
\newcommand{\cmark}{\ding{51}} 
\newcommand{\xmark}{\ding{55}} 
\usepackage{placeins}
\usepackage{tikz}
\usetikzlibrary{calc}
\usetikzlibrary{tikzmark}
\usepackage[accsupp]{axessibility}  
\usepackage{arydshln}

\definecolor{iccvblue}{rgb}{0.21,0.49,0.74}
\usepackage[pagebackref,breaklinks,colorlinks,allcolors=iccvblue]{hyperref}

\def\paperID{*****} 
\def\confName{ICCV}
\def\confYear{2025}

\title{Interpretable Decision-Making for End-to-End Autonomous Driving}

\author{Mona Mirzaie \quad Bodo Rosenhahn\\
Institute for Information Processing, Leibniz University Hannover\\
{\tt\small mirzaie@tnt.uni-hannover.de}
}

\begin{document}
\maketitle
\begin{abstract}
Trustworthy AI is mandatory for the broad deployment of autonomous vehicles. Although end-to-end approaches derive control commands directly from raw data, interpreting these decisions remains challenging, especially in complex urban scenarios. This is mainly attributed to very deep neural networks with non-linear decision boundaries, making it challenging to grasp the logic behind AI-driven decisions. This paper presents a method to enhance interpretability while optimizing control commands in autonomous driving. To address this, we propose loss functions that promote the interpretability of our model by generating sparse and localized feature maps. The feature activations allow us to explain which image regions contribute to the predicted control command. We conduct comprehensive ablation studies on the feature extraction step and validate our method on the CARLA benchmarks. We also demonstrate that our approach improves interpretability, which correlates with reducing infractions, yielding a safer, high-performance driving model. Notably, our monocular, non-ensemble model surpasses the top-performing approaches from the CARLA Leaderboard by achieving lower infraction scores and the highest route completion rate, all while ensuring interpretability. Code is available \href{https://github.com/MandM-VisionLab/DTCP}{here}\footnote{\href{https://github.com/MandM-VisionLab/DTCP}{https://github.com/MandM-VisionLab/DTCP}}.
\end{abstract}

\section{Introduction}
\label{sec:intro}
Autonomous driving systems in use today are categorized as either highly modularized hand-engineered, or end-to-end approaches. The modular pipeline is the conventional method widely adopted in the industry, which consists of several subsystems like perception, localization, prediction, planning, and control \cite{levinson2011towards}. The main advantage of the modular pipeline lies in its interpretability, which allows the pinpointing of malfunctioning modules \cite{tampuu2020survey, wang2021versatile}. Nevertheless, a closer look at the perception module shows that even with the integration of certain learning methods, its design still heavily relies on human-defined features, such as lane markings and vehicles \cite{kim2017interpretable}. The reliance on human heuristics (e.g.\hspace{0.3em}rule‑based planners) can result in overly conservative driving policies and difficulties in generalization, as each new driving scenario might require reconfigured heuristics \cite{chen2021interpretable, wang2021versatile}. Furthermore, the development and maintenance of this system are expensive and overly complex \cite{yurtsever2020survey, chen2022milestones}. Despite years of sustained efforts, these approaches still fall short of full autonomy \cite{tampuu2020survey, chen2021interpretable}.
\newline\indent Recent advancements in the realm of autonomous driving have shown great potential for end-to-end approaches. Human drivers often rely on reflex-like actions, requiring minimal high-level reasoning and conscious attention—a concept aligned with end-to-end learning \cite{tampuu2020survey}. Moreover, the limitations of the conventional methods can be mitigated through end-to-end autonomous driving techniques, where a driving policy is learned and generalized to new tasks with minimal manual engineering \cite{chen2021interpretable, wang2021versatile}. While ensuring trustworthiness in autonomous driving necessitates interpretability to address safety concerns and support commercialization \cite{kuznietsov2024explainable}, achieving this remains a major challenge in end-to-end systems \cite{yurtsever2020survey, kim2017interpretable, zeng2019end, chen2024end}. The absence of intermediate outputs results in a black-box learning scheme \cite{mueller2018driving, zablocki2022explainability, xu2024drivegpt4}, which makes it difficult to identify the origins of errors and to clarify the rationale behind certain driving decisions  \cite{zeng2019end,xiao2020multimodal,kim2017interpretable}. To alleviate this limitation, we have incorporated a technique into our model that enables the acquisition of diverse feature maps.  This enhancement significantly improves interpretability within our model, leading to a notable reduction in infractions during driving sessions in the CARLA driving simulator \cite{dosovitskiy2017carla}. Several top-ranked approaches on the CARLA public Leaderboard \cite{carla_leaderboard} fuse multiple modalities, such as LiDAR and multi-view camera inputs, and leverage ensemble and knowledge distillation techniques, as well as incorporate many auxiliary tasks. Conversely, our work attains state-of-the-art performance by employing solely a monocular camera, without the aid of ensemble or distillation techniques and traffic-rule-specific sub-tasks. It also offers substantial reductions in execution time and computational complexity compared to other competitors on the CARLA Leaderboard. Our main contributions are as follows:
\begin{itemize}
    \item We present a mechanism that improves interpretability by enforcing diversity in learned features.
    \item We evaluate our model’s interpretability and show significant improvements over the baseline, TCP \cite{wu2022trajectory}.
    \item We conduct an in-depth empirical analysis on the LAV benchmark and validate our approach on the official CARLA Leaderboard.  
\end{itemize}
\section{Related Work}
\label{sec:re-works}
\textbf{Visual Explanation in Autonomous Driving:} Interpreting a model’s decisions is crucial for diagnosing failures and enhancing system performance. Methods to improve interpretability generally fall into five categories: attention-based mechanisms, utilization of semantic inputs, auxiliary tasks, cost learning, and the application of natural language processing (NLP) \cite{chen2024end, zablocki2022explainability}. 
\newline \indent Attention-based models reveal internal reasoning by highlighting salient regions \cite{kim2017interpretable}, or leveraging object-level features \cite{wang2019deep}. In parallel, auxiliary-task strategies leverage multi-task optimization to decode latent features into other relevant outputs—such as classification \cite{chekroun2023gri, toromanoff2020end, shao2023safety}, semantic segmentation \cite{huang2020multi,chitta2022transfuser, xiao2020multimodal, chekroun2023gri, chen2022learning}, depth estimation \cite{natan2022end, hawke2020urban, ishihara2021multi}, affordance predictions \cite{toromanoff2020end, mehta2018learning}, or motion prediction \cite{shao2023safety, chen2022learning}—thus promoting interpretable and robust representations. Semantic inputs offer structured high-level context by feeding e. g. bird’s-eye-view (BEV) representations \cite{bansal2018chauffeurnet, kim2020attentional} or rasterized scenes \cite{djuric2020uncertainty} directly to the network and hence enhancing interpretability through inherently understandable representations. Finally, cost-volume visualization (cost maps) provides transparent insights into the model’s decision-making by ranking sampled trajectories based on the predicted cost for reaching specified locations over time \cite{drews2017aggressive, falcone2007predictive, zeng2019end}. By displaying these maps for multiple future steps, the system’s trajectory choices and underlying rationales become more interpretable.
\newline \indent \textbf{Interpretability in Other Applications:} Recent advancements in fine-grained image classification have led to significant improvements in interpretability. B-Cos Networks \cite{bohle2022b} enhance interpretability by aligning input features with model weights, while ProtoPNet \cite{chen2019looks} and BotCL \cite{wang2023learning} achieve this through prototype-based reasoning and refined concept-based learning. Norrenbrock \etal \cite{norrenbrock2022take} and Chang \etal \cite{chang2020devil} develop lightweight training techniques that effectively discern both distinctive and exclusive features, enabling fine-grained class differentiation without complex architectures. Collectively, these works underscore the role of representation learning in enhancing interpretability by ensuring models learn distinct, meaningful features. Motivated by \cite{norrenbrock2022take,chang2020devil}, we design a novel feature diversity loss and integrate it into our training process alongside other losses, leading to more refined and localized feature maps. Our approach results in a more interpretable model, as demonstrated in \cref{subsec:ablation_study}. Our autonomous driving framework, which directly enhances interpretability in its model design, is a combination of cost learning and attention-like mechanisms. As the feature diversity loss acts as a regularization term, it penalizes redundant or overlapping feature activations while enforcing diversity in learned representations. Although our framework does not explicitly implement an attention mechanism like those in transformers or self-attention layers, the sparse feature selection process mimics the behavior of attention. It filters out irrelevant information and attends to the most relevant features leading to localized and interpretable feature selection, which is characteristic of attention models that focus on salient regions. 
\newline \indent \textbf{Evaluating Interpretability:} Recent models integrate motion prediction and planning within a single network, leveraging these components to enhance interpretability. For instance, PlanT \cite{renz2023plant} proposes a Relative Filtered Driving Score (RFDS) to quantify how well the planner focuses on the most critical objects for safe driving decisions, while UniAD \cite{hu2023planning} employs cost-map visualizations to highlight the safety risks leading to potential collisions across different trajectories.
Beyond these approaches, models incorporating transformers and attention mechanisms, such as TransFuser \cite{chitta2022transfuser}, ReasonNet \cite{shao2023reasonnet}, and InterFuser \cite{shao2023safety}, visualize attention weights directly to highlight the most influential tokens in decision-making. Comparing saliency maps with chosen ground-truth references, such as semantic concepts or bounding boxes, is another widely used technique, as demonstrated in \cite{zhang2018visual, zhou2019comparing}. Following Boggust \etal \cite{boggust2022shared}, we report Intersection over Union (IoU), Ground Truth Coverage (GTC), and Saliency Coverage (SC) metrics to quantify our model’s interpretability improvements, complemented by visualization results. To gauge how tightly our model’s visual attention aligns with its driving decisions, we compute the Pearson correlation between saliency mass and the control signals. Our results on the CARLA benchmarks show a substantial decrease in the infraction scores, further validating the effectiveness of our approach and highlighting the crucial role of interpretability in driving performance.  
\section{Method}
\label{sec:method}
In this section, we introduce our proposed diversity losses and how agents can benefit from using them during training to improve interpretability.
\subsection{Problem Setting}
\indent We optimize a driving policy $\pi$ that generates raw control signals $a = \{\text{brake}, \text{throttle}, \text{steer}\}$ in the ranges $[0,1]$, $[0,1]$, and $[-1,1]$ respectively. The inputs to $\pi$ include the sensor signal $s$, the vehicle speed $v$, and high-level navigation information $g$ (i.e., a discrete navigation command and the destination coordinates). To tackle this problem, we have focused on the Behavior Cloning (BC) approach in imitation learning (IL), which is a supervised learning method that pursues a policy $\pi$ to mimic the behavior of an expert $\pi ^ *$ \cite{wu2022trajectory, prakash2021multi, renz2023plant}. It can be formulated as: 
\begin{equation}
\scalebox{0.9}{$
\arg \min _\pi \mathbb{E}_{\left(\mathcal{X}, \mathrm{a}^*, \mathcal{W}\right) \sim \mathrm{D}}\left[\mathcal{L}_{\text{act}}\left(\mathrm{a}^*,\pi_{\text{act}}(\mathcal{X})\right) + \mathcal{L}_{\text{traj}}\left(\mathcal{W}, \pi_{\text{traj}}(\mathcal{X})\right)\right]
$}
\end{equation}
where the dataset $D$ is comprised of a sequence of state-action-trajectory tuples $\left\{(\mathcal{X},a^*,\mathcal{W})\right\}_{t=0}^T$ acquired from the expert by controlling the ego-vehicle and interacting with the environment in $T$ steps. The $\mathcal{L}_{\text{act}}$ loss function measures the deviation between expert action $a^*$ and the model's predicted action $\pi_{\text{act}}(\mathcal{X})$, whereas $\mathcal{L}_{\text{traj}}$ aims to minimize the distance between the ground truth trajectory $\mathcal{W}$ and the model's predicted trajectory $\pi_{\text{traj}}(\mathcal{X})$.
\subsection{Baseline} \label{subsec:baseline}
\subsubsection{End-to-end Driving Model} \label{subsec:model_arch}
\begin{figure*}[t]
  \centering
   \includegraphics[width=0.75\textwidth]{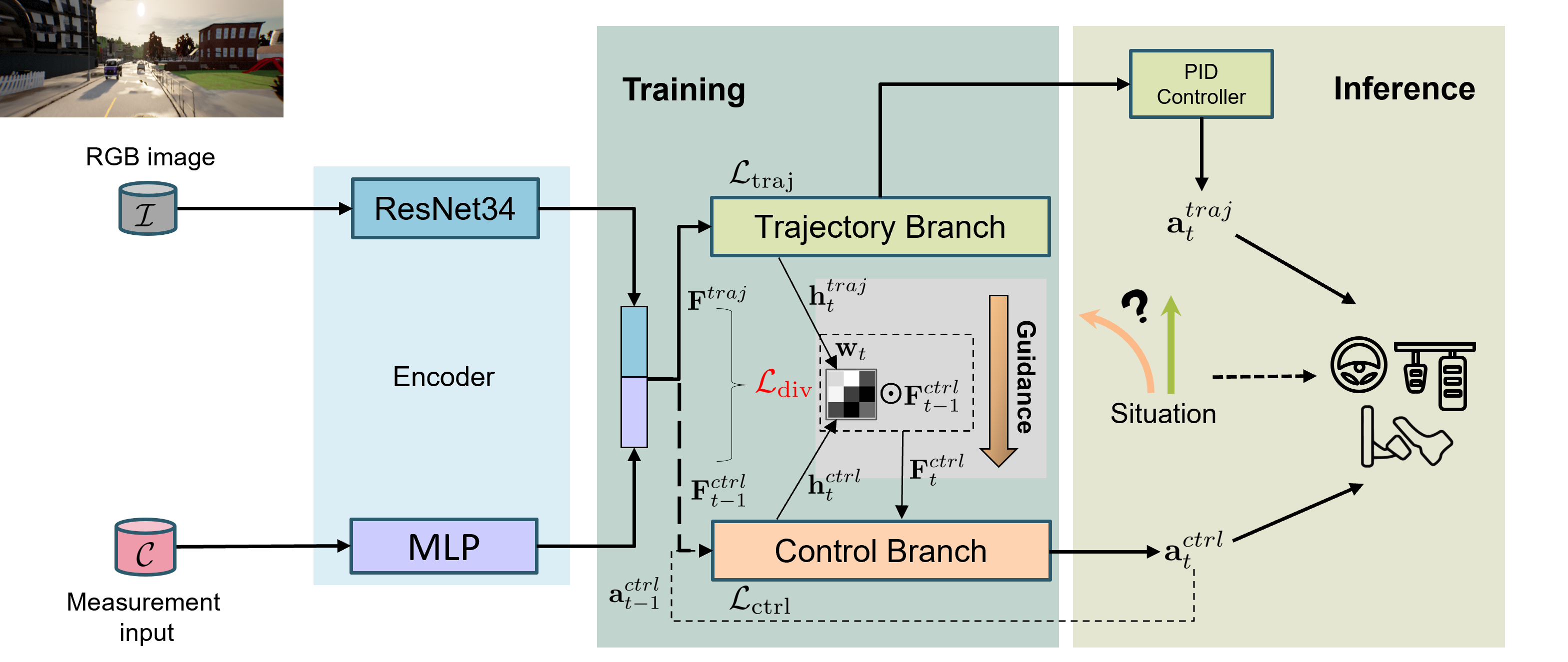}
\caption{\textbf{End-to-End Driving Framework:} The TCP (Trajectory-guided Control Prediction) predicts control signals $\mathbf{a}^{traj,ctrl} $ using a front-camera image $\mathcal{I}$ and a set of measurement data $\mathcal{C}$ (navigational commands, velocity, target point). $\mathbf{F}^{traj}$ denotes concatenated image and measurement encoder features, whereas $\mathbf{F}_{t-1}^{ctrl}$ represents measurement features fused with attention-weighted image features at the current step $t-1$ (dashed arrow). In the trajectory unit, waypoints are predicted using a GRU layer, while the control branch forecasts multi-step control actions leveraging the trajectory branch. During training, two widely adopted loss functions, $\mathcal{L}_{\mathrm{traj}}$ and $\mathcal{L}_{\mathrm{ctrl}}$, together with our proposed $\mathcal{L}_{\mathrm{div}}$ are applied to minimize the difference between the predicted waypoints and actions, and those provided by the expert. During inference, the converted control signals from a PID controller $\mathbf{a}^{traj}$ and the control branch $\mathbf{a}^{ctrl}$ are aggregated using the situational action fusion to form the final control actions.}
   \label{fig:tcp}
\end{figure*}

We adopt TCP (Trajectory-guided Control Prediction) \cite{wu2022trajectory} as our base model for its strong performance on the public CARLA Leaderboard 1.0 \cite{carla_leaderboard} and minimal reliance on extra modules. Unlike many imitation learning approaches that employ a rule-based expert agent, TCP utilizes Roach \cite{zhang2021end}, an expert agent based on Reinforcement Learning, offering the advantage of being trainable. TCP comprises trajectory and control branches to forecast the planned trajectories and multi time-steps control signals. In the TCP architecture presented in \cref{fig:tcp}, the front camera image $\mathcal{I}$ is processed by a pre-trained ResNet-34 encoder. Concurrently, the measurement input $\mathcal{C}$, which includes navigational commands and the goal location, along with the velocity is fed into an MLP encoder. The concatenated feature map from both encoders is then utilized by the control and trajectory prediction branches. Lastly, the predicted actions from each branch, i.e., $\mathbf{a}^{traj}$ and $\mathbf{a}^{ctrl}$, will be combined using a situation-based fusion strategy to form the final action for controlling the agent. In the following, each part will be discussed succinctly. For additional details, we refer to \cite{wu2022trajectory}.
\newline \indent \textbf{Trajectory Branch:} Motivated by the auto-regressive waypoint prediction network in \cite{prakash2021multi}, the concatenated feature map from both encoders $\mathbf{F}^{traj}$, along with the current position of the ego agent and the goal location are fed into GRU \cite{cho2014learning} layer, followed by linear forecasting of $T$ future time-step waypoints $\left\{\mathbf{w}_t\right\}_{t=1}^T$. This prediction involves computing the distance between each pair of future waypoints $\left\{\Delta \mathbf{w}_t\right\}_{t=1}^T$ and adding them to the current position $\left\{\mathbf{w}_{t-1}\right\}_{t=1}^T$ at each time-step, starting from the initial position $(0,0)$. 
\newline \indent To predict control actions $\mathbf{a}^{traj}$ from the forecasted future waypoints, an inverse dynamics model \cite{bellman1959adaptive} is implemented in the form of a PID controller, as presented in \cite{chen2020learning}. Particularly, two PID controllers are deployed for lateral and longitudinal control, taking the predicted future waypoints as input. The longitudinal controller adjusts throttle and brake, while the lateral one controls the steering.
\indent \textbf{Control Branch:} For forecasting multiple actions towards the future, a multi-step control strategy is applied. Given the current state $\mathcal{X}_{t-1}$, the control branch predicts future actions $\left\{\mathbf{a}^{ctrl}_{t-1}\right\}_{t=1}^T$ for $T$ time-steps based on the policy $\pi_{\text {ctrl}}\scriptstyle(\mathcal{X}_{t-1})$ using a temporal module, which delivers information about interaction between the ego-agent and the environment. This module uses a GRU layer that takes the feature at the current time step $\mathbf{F}_{t-1}^{ctrl}$ concatenated with the current predicted action $\mathbf{a}^{ctrl}_{t-1}$ as input and updates the hidden state $\mathbf{h}_{t-1}^{ctrl}$ for the next time-step $t$. Afterwards, an attention map $\mathbf{w}_{t}$ is computed using the hidden state from the GRU layer in the trajectory branch $\mathbf{h}_{t}^{traj}$ and the corresponding one in the temporal module $\mathbf{h}_{t}^{ctrl}$. The attention weights are multiplied element-wise with the feature map and then combined with the hidden state of the temporal module $\mathbf{h}_{t}^{ctrl}$. Accordingly, a policy head predicts the control action for the next time-step $t$ using the updated feature map $\mathbf{F}_{t}^{ctrl}$. Note that, successive GRU layer forward passes receive the hidden state of the previous time-step as the initial hidden state.
\newline  \indent \textbf{Control Actions Fusion:} During inference, the final driving action $\mathcal{A}$ is a weighted fusion of $\mathbf{a}^{traj}$ and $\mathbf{a}^{ctrl}$. During turning, $\mathbf{a}^{ctrl}$ receives a higher weight to reduce infractions, as discussed in \cite{wu2022trajectory}.
\vspace{0.5em}
\subsubsection{Baseline losses}  \label{subsec:baseline_losses}
TCP employs common baseline losses for training, including waypoint loss $\mathcal{L}_{\mathrm{traj}}$ \cite{chen2020learning, chitta2022transfuser, zhang2024feedback}, control action loss $\mathcal{L}_{\mathrm{ctrl}}$ \cite{jia2023driveadapter, zhang2021end, jia2023think, zhao2021sam}, and sub-task loss $\mathcal{L}_{\mathrm{sub}}$ \cite{zhang2021end, zhang2024feedback, codevilla2019exploring}. Each loss term is weighted by $\lambda$ to balance its contribution:
\begin{equation}
\mathcal{L}_{\mathrm{baseline}}= \lambda_{\mathrm{traj }} \cdot \mathcal{L}_{\mathrm{traj}}+\lambda_{\mathrm{ctrl}} \cdot \mathcal{L}_{\mathrm{ctrl}}+\lambda_{\mathrm{sub}} \cdot \mathcal{L}_{\mathrm{sub}}
\label{eq:baseline_losses}
\end{equation}
In the following, we briefly describe each loss term.
\newline \indent \textbf{Trajectory Loss:} The trajectory loss decreases the $L_1$ distance between the predicted waypoints $\mathbf{w}$ and the expert (ground truth) waypoints $\mathbf{\tilde{w}}$ for $T$ time-steps:
\begin{equation}
\scalebox{0.98}{$
\mathcal{L}_{\text {traj}}=\sum_{t=1}^T\left\|\mathbf{w}_t-\mathbf{\tilde{w}}_t\right\|_1
$}
\end{equation} 
\vspace{-1em}
\newline \indent \textbf{Action Loss:} Following \cite{zhang2021end}, predicted actions from the control unit $\mathbf{a}^{ctrl}$ and the corresponding expert ones $\tilde{\mathbf{a}}$ are transformed into Beta distributions $\mathcal{B}$. The KL-divergence between these distributions is minimized over current and future time-steps \cite{wu2022trajectory}:
\begin{equation}
\scalebox{0.87}{$
\mathcal{L}_{\mathrm{ctrl}}=\mathbf{KL}\left(\operatorname{\mathcal{B}}\left(\mathbf{a}^{ctrl}_0\right) \| \mathcal{B}\left(\tilde{\mathbf{a}}_0\right)\right)+\frac{1}{T} \sum_{t=1}^T \mathbf{K L}\left(\mathcal{B}\left(\mathbf{a}^{ctrl}_t\right) \| \mathcal{B}\left(\tilde{\mathbf{a}}_t\right)\right)
\label{eq:action_loss}
$}
\end{equation}
\newline \indent \textbf{Sub-tasks Losses:} As in \cite{zhang2021end}, we include auxiliary tasks—feature matching ($\mathcal{L}_{\mathrm{F}}$), speed regulation ($\mathcal{L}_{\mathrm{S}}$), and value regression ($\mathcal{L}_{\mathrm{V}}$)—to align model predictions closer to expert behavior. Specifically, the feature matching loss $\mathcal{L}_{\mathrm{F}}$ enforces consistency between latent representations of the model and the expert, while the speed regulation loss $\mathcal{L}_{\mathrm{S}}$ minimizes differences between the predicted and expert-desired speeds. The value regression loss $\mathcal{L}_{\mathrm{V}}$ predicts the expected future return, hinting at possible upcoming hazardous incidents. These losses are combined as follows:
\begin{equation}
\mathcal{L}_{\mathrm{sub}}=\lambda_{\mathrm{F}}\mathcal{L}_{\mathrm{F}}+\lambda_{\mathrm{S}}\mathcal{L}_{\mathrm{S}}+\lambda_{\mathrm{V}}\mathcal{L}_{\mathrm{V}}
\end{equation}
\subsection{Our Loss Design} \label{subsec:loss}
The training loss of our proposed method introduces a novel diversity term  $\mathcal{L}_{\mathrm{div}}$, alongside baseline losses (see \cref{eq:baseline_losses}), expressed as:
\begin{equation}
\mathcal{L}=\lambda_{\mathrm{div }} \cdot \mathcal{L}_{\mathrm{div }}+\lambda_{\mathrm{baseline}} \cdot \mathcal{L}_{\mathrm{baseline}}
\label{eq:all_losses_compact}
\end{equation} 
\newline \indent \textbf{Diversity Loss:} Inspired by \cite{norrenbrock2022take}, we propose to incorporate a diversity loss during training, which encourages the model to capture more distinct features that are distributed as widely as possible. Our proposed loss is highlighted in red in \cref{fig:tcp}.  For a specific feature map $l \in\left\{0,1, \ldots, n_f-1\right\}$ within the feature maps $M \in \mathbb{R}^{n_f \times w_M \times h_M}$, the diversity component is formulated as:
\begin{equation}
\hat{s}_{i j}^l=\frac{\exp \left(m_{i j}^l\right)}{\sum_{i^{\prime}=1}^{h_M} \sum_{j^{\prime}=1}^{w_M} \exp \left(m_{i^{\prime} j^{\prime}}^l\right)} \frac{\bar{\boldsymbol{f}_l}}{\max _{c\in\left[1, {n_f}\right]} \bar{\boldsymbol{f}_c}} 
\end{equation}
For each activation value $m_{ij} $ in the ${l}^{th}$ feature map of $M$, we apply a softmax normalization across the spatial dimensions ${h_M} \times {w_M}$, followed by scaling based on the ratio of the ${l}^{th}$ feature map's average activation $\bar{f_l}$ to the maximum average activation $\bar{f_c}$ across all ${n_f}$ feature maps. Accordingly, the diversity loss $\mathcal{L}_{\mathrm{div}}$ aims to minimize the sum of the maximum activations, meaning that the weighted feature map $\hat{S}^l$ highlights different regions of the input. This yields a rich variety of relevant feature activations, defined as:
\begin{equation}
\scalebox{0.99}{$
\mathcal{L}_{\mathrm{div}}=-\sum_{i=1}^{h_M} \sum_{j=1}^{w_M} \max \left(\hat{s}_{i j}^1, \hat{s}_{i j}^2, \ldots, \hat{s}_{i j}^{n_f}\right)
$}
\end{equation}
This approach prevents redundancy and ensures the network exploits more diverse features across the feature maps. The diversity loss (see \cref{eq:div_loss_final}) in our approach is applied to both trajectory and control branches. For the trajectory unit, we consider only the current step feature $\mathbf{F}^{traj}$ containing the concatenated feature map from both trajectory and measurement encoders. In contrast, for the control branch, we average the loss of $T$ future time-steps $\left\{\mathbf{F}_{t}^{ctrl}\right\}_{t=1}^T$. This strategy prioritizes the optimization of the actual control action, thereby underscoring the importance of feature diversity at the current time-step.
\begin{equation}
\resizebox{\linewidth}{!}{$
\displaystyle
\mathcal{L}_{\mathrm{div}}^{branches} = \lambda_{\mathrm{div}} \cdot \biggl(
\mathcal{L}_{\mathrm{div}}^{traj}\left(\mathbf{F}_{0}^{traj}\right) 
+ \mathcal{L}_{\mathrm{div}}^{ctrl}\left(\mathbf{F}_{0}^{ctrl}\right)
+ \frac{1}{T} \sum_{t=1}^T\mathcal{L}_{\mathrm{div}}^{ctrl}\left(\mathbf{F}^{ctrl}\right)\biggr)
$}
\label{eq:div_loss_final}
\end{equation}
\section{Experiments}
\label{sec:exps}
\subsection{Experiment Setup} \label{subsec:exps_setup}
We conduct our experiments using CARLA 0.9.10.1, comprising 8 publicly available towns \cite{dosovitskiy2017carla}.
\newline \indent \textbf{Dataset:} Following \cite{chen2022learning, wu2022trajectory}, we use the public TCP dataset containing 420\,k  frames from 8 cities collected using the Roach driving expert \cite{zhang2021end} under randomized weather. For the ablation study (\cref{subsec:ablation_study}),  we train on 185K frames from 4 towns and test on LAV validation routes  \cite{chen2022learning}, including two unseen towns. To further validate our approach, we evaluate the Transfuser model \cite{prakash2021multi} on the Transfuser dataset \cite{transfuser_dataset} with our method. For the leaderboard submission (\cref{subsec:benchmark_leaderboard}), we train on the full TCP dataset.
\newline \indent \textbf{Metrics:} Driving performance is assessed using three metrics \cite{carla_leaderboard}: Driving Score (DS), Route Completion (RC), and Infraction Penalty (IP). RC quantifies the percentage of the route successfully driven by the agent. IP penalizes violations of traffic rules starting from an ideal score of 1.0, decreasing according to the severity and frequency of infractions. DS summarizes the agent's performance, balancing route completion and safety, and is derived by weighting RC by infractions incurred. Infractions normalized per kilometer are also reported.
\newline \indent \textbf{Baselines:}  To benchmark our approach on the CARLA Leaderboard, we compare against four leading methods alongside TCP \cite{wu2022trajectory}. \textbf{InterFuser} \cite{shao2023safety} employs a transformer-based fusion of multi-view cameras and LiDAR data, predicting waypoints, object density maps, and traffic rules. It computes desired velocity using density maps combined with motion forecasts derived from historical vehicle dynamics. \textbf{ReasonNet} \cite{shao2023reasonnet} extends InterFuser by adding temporal reasoning modules for improved motion prediction and uses consistency loss to align occupancy maps with predicted trajectories. \textbf{TF++ WP} \cite{jaeger2023hidden} fuses front-camera images and LiDAR BEV representations, predicting future trajectories with auxiliary segmentation and detection tasks, controlled by a modified PID controller. \textbf{LAV} \cite{chen2022learning} integrates multi-view images and LiDAR into BEV representations, predicting semantic maps and dynamic object bounding boxes. It utilizes a privileged motion planner with a distillation framework, enabling the ego-agent to plan trajectories based solely on its own sensors, controlled by PID controllers coupled with collision prediction.
\newline \indent For our ablation study, in addition to TCP, we also evaluate \textbf{TransFuser} \cite{prakash2021multi}. It has a similar architecture to TF++ WP that uses transformers to fuse sensory inputs, including a LiDAR BEV and a front-view camera image, without using any auxiliary tasks, and predicts only the ego-vehicle's future waypoints using the trajectory loss (see \cref{subsec:baseline_losses}). During inference, two PID controllers process the predicted waypoints for lateral and longitudinal control. 
\newline \indent \textbf{Benchmarks:} We evaluate our model on two benchmarks: LAV routes and the CARLA public Leaderboard. The LAV benchmark \cite{chen2022learning} includes 16 route types in Towns 02 and 05 across 4 weather conditions and random scenarios from the CARLA Leaderboard. The CARLA public Leaderboard 1.0 \cite{carla_leaderboard} consists of 100 secret routes in unseen towns under adversarial conditions, following the NHTSA pre-crash typology \cite{carla_leaderboard, najm2007pre}. 
\newline \indent \textbf{Implementation Details:} We build our approach, named \textbf{DTCP} (Diversity-enhanced TCP), upon the TCP baseline~\cite{wu2022trajectory} and follow its supplementary material for hyperparameter settings and baseline loss weighting factors $\lambda$. Details regarding our method's hyperparameters such as $\lambda_{\mathrm{div}}$ can be found in our supplementary material. We set the future time-steps $T=4$ in \cref{eq:div_loss_final}, consistent with the choice in \cite{wu2022trajectory} for \cref{eq:action_loss}. Additionally, we reproduce TransFuser \cite{prakash2021multi} as reported in \cite{prakash2021supplementary} for our ablation study. This earliest version is computationally lighter than later versions, including TransFuser \cite{chitta2022transfuser} and TF++ WP \cite{jaeger2023hidden}.
\newline \indent \textbf{Model Validation:} Optimal model selection requires validation metrics that comprehensively capture diverse driving demands, a challenge often overlooked in CARLA benchmarks. Methods like Transfuser \cite{prakash2021multi} and TCP \cite{wu2022trajectory} prioritize trajectory accuracy but fail to assess robustness in complex scenarios (e.g., traffic avoidance, road sign compliance). To address this, we evaluate our top three checkpoints directly in CARLA on the validation routes and select the best-performing model for benchmarking.
\subsection{Ablation Study} \label{subsec:ablation_study}
\textbf{Loss Component Analysis:} We report our results, which are the mean and standard deviation (std) of five evaluation runs, on LAV benchmark \cite{chen2022learning} with two different baselines demonstrating the effectiveness of feature diversity in driving performance shown in \cref{tab:benchmark_lav_DS}.
\begin{table}[htbp]
 \caption{\textbf{Results on LAV Benchmarks}. We compare the integration of our proposed losses within two baselines, TCP and TransFuser, demonstrating enhanced driving performance and infraction score. Results are averaged over five test runs. The value band \emph{Expert} indicates an upper bound for our IL agent.}
  \label{tab:benchmark_lav_DS}
  \centering
  \resizebox{\columnwidth}{!}{
  \begin{tabular}{
    l
    >{\centering\arraybackslash}p{2.5cm}
    >{\centering\arraybackslash}p{2.5cm}
    >{\centering\arraybackslash}p{2.5cm}
    }
    \toprule
    {Method} &
    \multicolumn{1}{c}{\makecell{Driving \\ score \\ $\%,\uparrow$}} &  
    \multicolumn{1}{c}{\makecell{Route \\ completion \\ $\%,\uparrow$}} & 
    \multicolumn{1}{c}{ \makecell{Infraction \\ penalty \\ $[0,1],\uparrow$}} \\
    \midrule
     \emph{Roach Expert} \cite{zhang2021end} & 77.96 $\pm$ 1.12 & 98.35 $\pm$ 1.73 & 0.79 $\pm$ 0.01  \\
    \addlinespace[0.2em] \cdashline{1-4} \addlinespace 
     TCP \cite{wu2022trajectory} & 53.10 $\pm$ 2.98 &  88.39 $\pm$ 3.14 & 0.60 $\pm$ 0.02 \\
     DTCP (\textbf{ours}) & \textbf{61.87} $\pm$ 1.35 & \textbf{98.35} $\pm$ 1.67 & \textbf{0.63} $\pm$ 0.01 \\ 
    \addlinespace
    \cline{1-4}
    \addlinespace
     \emph{Rule-based Expert} & 68.52 $\pm$ 1.15 & 97.98 $\pm$ 0.31 & 0.69 $\pm$ 0.01 \\ 
    \addlinespace[0.2em] \cdashline{1-4} \addlinespace 
     TransFuser \cite{prakash2021multi} & 19.71 $\pm$ 0.79 & 48.09 $\pm$ 4.98 & 0.56 $\pm$ 0.06 \\
     DTCP (\textbf{ours}) & \textbf{29.60} $\pm$ 0.87 & \textbf{55.84} $\pm$ 6.80 & \textbf{0.58} $\pm$ 0.08 \\ 
    \bottomrule
  \end{tabular}
  }
\end{table}
\begin{table*}[!htbp]
  \caption{\textbf{Infraction Details on LAV Benchmark}. We present a more detailed analysis of infractions across five runs. Our method significantly reduces nearly all types of collisions, matching expert-level performance in some categories.}
  \label{tab:lav_bench_infracs}
  \centering
  \resizebox{0.9\textwidth}{!}{
  \small 
  \begin{tabular}{
    @{}
    >{\centering\arraybackslash}p{2.5cm}
    l
    >{\centering\arraybackslash}p{1.5cm}
    >{\centering\arraybackslash}p{1.5cm}
    >{\centering\arraybackslash}p{1.5cm}
    >{\centering\arraybackslash}p{1.5cm}
    >{\centering\arraybackslash}p{1.5cm}
    >{\centering\arraybackslash}p{1.5cm}
    >{\centering\arraybackslash}p{1.5cm}
    >{\centering\arraybackslash}p{1.5cm}
    >{\centering\arraybackslash}p{1.5cm}
    @{}
  }
    \toprule
    Benchmark &
    \multicolumn{1}{l}{Method} &  
    \multicolumn{1}{c}{\makecell{Collisions\\pedestrians\\$\text{\#km},\downarrow$}} &  \multicolumn{1}{c}{\makecell{Collisions\\vehicles\\$\text{\#km},\downarrow$}} & \multicolumn{1}{c}{\makecell{Collisions\\layout\\$\text{\#km},\downarrow$}} & \multicolumn{1}{c}{\makecell{Red light\\infractions\\$\text{\#km},\downarrow$}} & \multicolumn{1}{c}{\makecell{Stop sign \\infractions\\$\text{\#km},\downarrow$}} & \multicolumn{1}{c}{\makecell{Off-road \\infractions\\$\text{\#km},\downarrow$}} & \multicolumn{1}{c}{\makecell{Route \\deviations\\$\text{\#km},\downarrow$}} & \multicolumn{1}{c}{\makecell{Route \\timeouts\\$\text{\#km},\downarrow$}}  & \multicolumn{1}{c}{\makecell{Agent \\blocked\\$\text{\#km},\downarrow$}} \\
        \midrule
        \centering
         \multirow{3}{*}{\makecell{LAV \\ Routes \\ \begin{minipage}{2cm}\centering\cite{chen2022learning}\end{minipage}}}
        & \emph{Roach Expert} \cite{zhang2021end} & 0.00 & 0.08 & 0.00 & 0.00 & 0.00 & 0.00 & 0.00 & 0.02 & 0.01 \\
         \addlinespace[0.2em]
        \cdashline{2-11} \addlinespace
        & TCP \cite{wu2022trajectory} & 0.00 & 0.09 & 0.04 & 0.04 & \bfseries 0.09 & 0.05 & 0.00 & 0.06 & 0.06 \\
        & DTCP (\textbf{ours}) & 0.00 & \bfseries 0.07 & \bfseries 0.01 & \bfseries 0.01 & 0.14 & \bfseries 0.02 & 0.00 & \bfseries 0.02 & \bfseries 0.01 \\ 
        \addlinespace[2pt]
        \bottomrule
      \end{tabular}
      }
    \end{table*}
\begin{figure*}[t]
\centering
\begin{minipage}[b]{0.3\textwidth}
  \centering
  \textbf{Input}\\[0.3em]
  \includegraphics[width=\textwidth]{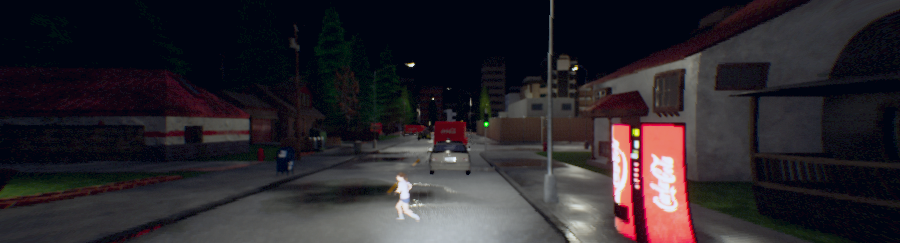}
\end{minipage}
\begin{minipage}[b]{0.3\textwidth}
  \centering
  \textbf{DTCP (ours)}\\[0.3em]
  \includegraphics[width=\textwidth]{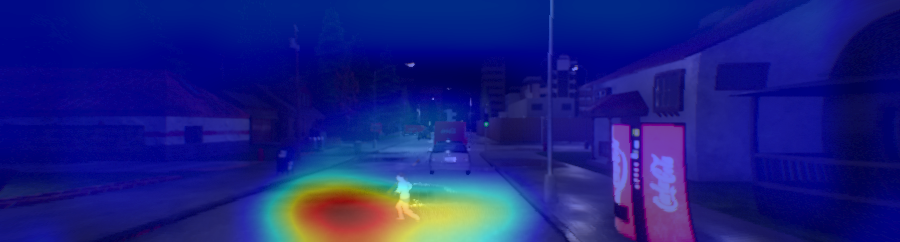}
\end{minipage}
\begin{minipage}[b]{0.3\textwidth}
  \centering
  \textbf{TCP}\\[0.3em]
  \includegraphics[width=\textwidth]{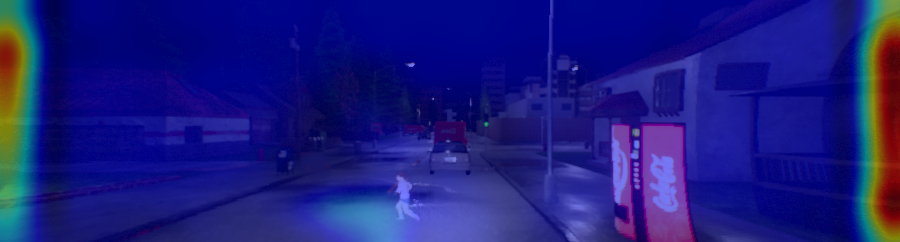}
\end{minipage}

\vspace{2pt} 



\begin{minipage}[b]{0.3\textwidth}
  \centering
  \includegraphics[width=\textwidth]{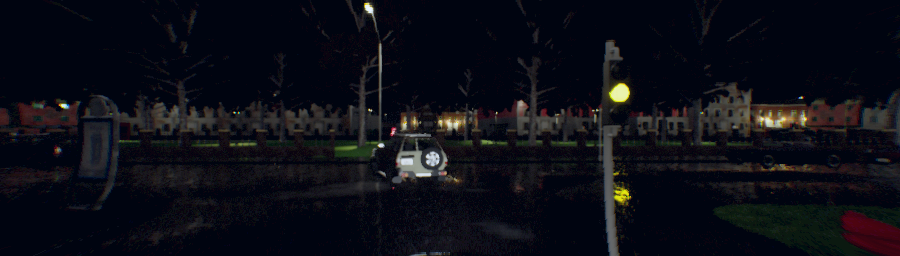}
\end{minipage}
\begin{minipage}[b]{0.3\textwidth}
  \centering
  \includegraphics[width=\textwidth]{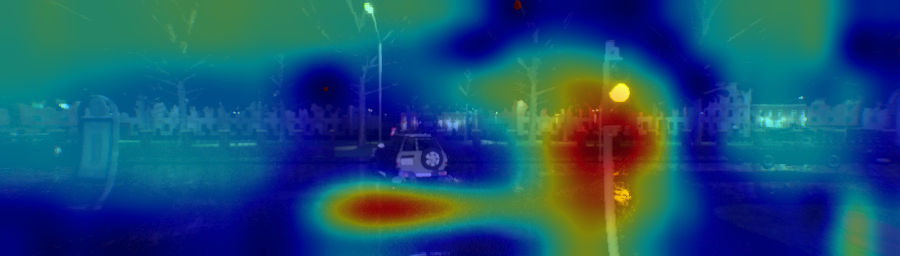}
\end{minipage}
\begin{minipage}[b]{0.3\textwidth}
  \centering
  \includegraphics[width=\textwidth]{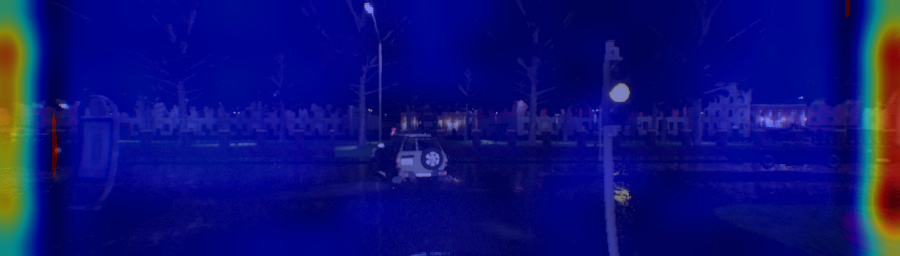}
\end{minipage}
  \caption{\textbf{EigenCam Visualizations in Various Challenging Scenarios}.
  From left to right: original image, DTCP (ours), and reproduced TCP. TCP shows a uniform feature distribution, primarily focusing on the left and right due to crossing traffic participants. In contrast, our method captures more diverse feature representations, enhancing focus on regions critical for driving decisions.}
  \label{fig:eigCam_vis}
\end{figure*}
\begin{table*}[!htbp]
 \caption{\textbf{Quantitative Evaluation of Interpretability for Infraction-Related Categories}. We report the mean for each metrics. Our method achieves the highest GTC, indicating that it allocates greater attention to objects of interest compared to the reproduced TCP. This result demonstrates that incorporating the diversity loss helps to meaningfully localize the feature map.}
  \label{tab:heatmap_compare}
  \centering
  \normalsize
  \resizebox{\textwidth}{!}{
  \begin{tabular}{
    >{\centering\arraybackslash}p{2cm}
    l
    >{\centering\arraybackslash}p{1.5cm} >{\centering\arraybackslash}p{1.5cm} >{\centering\arraybackslash}p{1.5cm}|
    >{\centering\arraybackslash}p{1.5cm} >{\centering\arraybackslash}p{1.5cm} >{\centering\arraybackslash}p{1.5cm}|
    >{\centering\arraybackslash}p{1.5cm} >{\centering\arraybackslash}p{1.5cm} >{\centering\arraybackslash}p{1.5cm}|
    >{\centering\arraybackslash}p{1.5cm} >{\centering\arraybackslash}p{1.5cm} >{\centering\arraybackslash}p{1.5cm}|
    >{\centering\arraybackslash}p{1.5cm} >{\centering\arraybackslash}p{1.5cm} >{\centering\arraybackslash}p{1.5cm}
    }
    \toprule
    \multirow{2}{*}{\textbf{Routes}} &
    \multirow{2}{*}{\textbf{Method}} &
    \multicolumn{3}{c|}{\textbf{Pedestrians}} &  
    \multicolumn{3}{c|}{\textbf{Cyclists}} &  
    \multicolumn{3}{c|}{\textbf{Vehicles}} &  
    \multicolumn{3}{c|}{\textbf{Traffic Lights}} &  
    \multicolumn{3}{c}{\textbf{Overall}} \\
    \cmidrule(lr){3-5} \cmidrule(lr){6-8} \cmidrule(lr){9-11} \cmidrule(lr){12-14} \cmidrule(lr){15-17}
    & & IoU & GTC & SC & IoU & GTC & SC & IoU & GTC & SC & IoU & GTC & SC & IoU & GTC & SC \\
    \midrule
    \multirow{2}{*}{Town01}
    & TCP \cite{wu2022trajectory} & 0.01 & 0.07 & 0.01 & 0.01 & 0.12 & 0.02 & 0.01 & 0.06 & 0.01 & \textbf{0.08} & 0.23 & \textbf{0.09} & 0.02 & 0.09 & 0.02\\
    & DTCP (\textbf{ours}) & 0.01 & \textbf{0.60} & 0.01 & \textbf{0.02} & \textbf{0.44} & 0.02 & 0.01 & \textbf{0.25} & \textbf{0.02} & 0.05 & \textbf{0.35} & 0.06 & 0.02 & \textbf{0.30} & 0.02\\ 
    \midrule
    \multirow{2}{*}{Town02}
    & TCP \cite{wu2022trajectory} & 0.00 & 0.00 & 0.00 & 0.01 & 0.11 & 0.01 & 0.01 & 0.04 & 0.01 & \textbf{0.07} & 0.26 & \textbf{0.08} & 0.02 & 0.08 & 0.02 \\
    & DTCP (\textbf{ours}) & \textbf{0.01} &  \textbf{0.45} &\textbf{0.01} & \textbf{0.03} & \textbf{0.55} & \textbf{0.03} & \textbf{0.02} & \textbf{0.33} & \textbf{0.02} & 0.05 & \textbf{0.28} & 0.06 & 0.02 & \textbf{0.33} & 0.02 \\ 
    \midrule
    \multirow{2}{*}{Town03}
    & TCP \cite{wu2022trajectory} & 0.00 & 0.00 & 0.00 & 0.00 & 0.02 & 0.00 & 0.01 & 0.09 & 0.01 & 0.00 & 0.00 & 0.00 & 0.01 & 0.06 & 0.01\\
    & DTCP (\textbf{ours}) &  \textbf{0.01} & \textbf{0.87} &  \textbf{0.01} & \textbf{0.01} & \textbf{0.40} & 0.00 & 0.01 & \textbf{0.19} & 0.01 & 0.00 & \textbf{0.70} & 0.00 & 0.01 & \textbf{0.28} & 0.01\\ 
    \midrule
    \multirow{2}{*}{Town05}
    & TCP \cite{wu2022trajectory} & 0.00 & 0.02 & 0.00 & 0.00 & 0.00 & 0.00 & 0.01 & 0.07 & 0.01 & 0.00 & 0.00 & 0.00 & 0.01 & 0.06 & 0.01\\
    & DTCP (\textbf{ours}) & \textbf{0.01} & \textbf{0.59} & \textbf{0.01} & 0.00 & \textbf{0.07} & 0.00 & 0.01 & \textbf{0.27} & 0.01 & 0.00 & \textbf{0.19} & 0.00 & 0.01 & \textbf{0.27} & 0.01\\ 
    \bottomrule
  \end{tabular}
  }
\end{table*}
\newline \indent Using our diversity loss in TCP significantly improves DS by 8\%, RC by 10\%, and IP by 0.03. We observe a reduction in nearly all infractions as depicted in \cref{tab:lav_bench_infracs}. Specifically, it reduces both red light violation and layout collisions by a factor of 4. 
\newline \indent Applying our approach to the TransFuser baseline increases the route completion by 7\% and driving score by 10\%. Our model maintains a better trade-off between safety and driving efficiency, a balance desired in every autonomous driving model. 
\newline \indent \textbf{Interpretability:} We visualize the last layer of the image encoder for both the reproduced TCP and our model in \cref{fig:eigCam_vis} using EigenCam \cite{muhammad2020eigen}, which projects latent features onto eigenvectors to highlight critical regions. Due to incorporating the feature diversity loss ($\mathcal{L}_{\mathrm{div}}$), interpretability improves compared to the reproduced version. Specifically, our model focuses on the crossing pedestrian in the first row, while also considering the risk of yellow traffic lights (last row) \cite{wu2022trajectory}.
\newline \indent We further validate the interpretability improvement using \emph{Shared Interest} metrics (IoU, GTC, SC) \cite{boggust2022shared}, comparing model-generated saliency maps to ground-truth bounding boxes of high-risk infraction categories (details in Supplementary). IoU measures the intersection of the ground truth set ($\mathcal{G}$) and the saliency set ($\mathcal{S}$) relative to their union ($\frac{I}{U}$). GTC indicates the proportion of ground truth features covered by the saliency set ($\frac{I}{\mathcal{G}}$), and SC denotes the proportion of saliency features matching the ground truth ($\frac{I}{\mathcal{S}}$). As shown in \cref{tab:heatmap_compare}, our model achieves substantially higher GTC across all categories than the reproduced TCP, indicating a stronger focus on ground‐truth features relevant for decision-making \cite{morrison2023shared,boggust2022shared}. Comparable IoU and SC scores between models are due to our model's reliance on diverse features beyond ground truth, and targets in each frame constitute a minor portion of the activation map. Additionally, not all bounding boxes directly impact decision-making, and partial overlaps can reduce metric scores despite appropriate saliency allocation (see Supplementary Material for visual examples). Our analysis shows that improved interpretability correlates with lower infraction rate, as prioritizing infraction-related objects enhances driving safety and performance, as corroborated by \cref{tab:benchmark_lav_DS}.
In our Supplementary Material, we additionally compute driving-related semantic IoU to demonstrate that DTCP assigns distinct attentions to semantically coherent regions, thereby supporting human-level interpretability. 
\newline \indent \cref{tab:corr_pearson} reports the Pearson correlation between saliency magnitude and steering signals derived from control actions $\mathbf{a}^{traj,ctrl}$. DTCP exhibits positive correlations, whereas the reproduced TCP shows negative correlations, resulting in substantially higher explained variance ($R^{2}$: $\sim${}10$\times$ \emph{traj}, $3\times$ \emph{ctrl}) and decisive Fisher $z$-scores ($z{=}5.6$, \emph{traj}; $7.8$, \emph{ctrl}). This further demonstrates that DTCP’s enhanced focus on driving-critical cues improves driving performance (cf. \cref{tab:benchmark_lav_DS}, \cref{tab:benchmark_leaderboard}).
\begin{table}[t]
\centering
\caption{\textbf{Saliency–Steering Pearson Correlation on LAV Routes.} $\rho$ denotes Pearson’s correlation between saliency mass and steering; $R^{2}{=}\rho_{\text{DTCP}}^{2}/\rho_{\text{TCP}}^{2}$ is the relative explained variance; $z$ is Fisher’s test statistic comparing correlations. DTCP correlations are \emph{positive}, whereas TCP correlations are \emph{negative}, indicating our model’s attention is directed toward driving-relevant cues.}
\resizebox{0.72\columnwidth}{!}{
\begin{tabular}{@{}lrrrr@{}}
\toprule
Control Signal & $\rho_{\text{DTCP}}$ & $\rho_{\text{TCP}}$ & $R^{2}$ & $z$ \\
\midrule
\textbf{steer-ctrl} & $+0.122$ & $-0.223$ & 0.30 & 7.84 \\
\textbf{steer-traj} & $+0.186$ & $-0.060$ & 9.76 & 5.57 \\
\bottomrule
\end{tabular}
}
\label{tab:corr_pearson}
\end{table}
\newline \indent \textbf{Limitations:} The results in \cref{tab:lav_bench_infracs} indicate that our method still struggles with stop signs. The reproduced TCP experiences route timeouts $3\times$ and agent blocks $6\times$ more often than DTCP, reflecting slower and hesitant driving. Stop sign infractions depend on slowing down to 2 m/s and stopping within the simulator-defined trigger area \cite{jaeger2023hidden}. Hence, the slower reproduced TCP meets this criterion more frequently than the faster DTCP, which achieves higher route completion. This discrepancy arises because the LAV validation routes predominantly feature road-painted 'STOP' markings rather than upright signs. While upright signs can be detected earlier, painted signs become visible for only a brief moment, making braking success primarily speed-dependent rather than perception-driven. Thus, reproduced TCP’s better performance here is incidental rather than a result of superior perception. For distant, upright signs, DTCP's diversified features provide better detection, explaining the reversed stop-sign infraction results observed between \cref{tab:lav_bench_infracs} and \cref{tab:leaderboard_bench_infracs}. Stop-sign infractions in our model can be substantially reduced by incorporating bounding-box detection, ensuring the controller initiates braking as soon as the vehicle enters the boundary of a painted stop sign, as suggested in \cite{jaeger2023hidden}. We further discuss failure cases in our Supplementary Material, illustrating particular limitations of using the \emph{Shared Interest} metrics (see \cref{tab:heatmap_compare}).
\subsection{State-of-the-Art Comparison} \label{subsec:benchmark_leaderboard}
We evaluate our approach on the public CARLA Leaderboard 1.0 \cite{carla_leaderboard}, based on CARLA version 0.9.10.1. Although the new version of leaderboard, Leaderboard 2.0, which is based on the CARLA version 0.9.14, is already available, we have decided to use Leaderboard 1.0 as it is more suitable for benchmarking due to its 31 submissions versus only 3 submissions in new version. We will update our results on the new leaderboard in future works. 
\newline \indent \textbf{CARLA Leaderboard 1.0:} \Cref{tab:benchmark_leaderboard} presents the results of our approach, denoted as DTCP, compared to the five top-ranked entries on the public CARLA Leaderboard 1.0 \cite{carla_leaderboard}. For a fair comparison, we also submitted a reproduced version of the TCP model. Conversely, the TCP model submitted by Wu \etal \cite{wu2022trajectory} comprises an ensemble model combination of two models TCP and TCP-SB. TCP-SB incorporates two distinct encoders for both branches instead of utilizing shared ones with the trajectory unit supervision (cf. \cref{fig:tcp}). Our model secured 4th place among all 31 leaderboard submissions with the highest route completion. It outperforms reproduced TCP by 12\% on both DS and RC with an improvement of 0.06 in IP. This demonstrates that despite utilizing only a monocular camera, our model exhibits enhanced object detection ability owing to its increased interpretability, although all other models using multi camera-views and a LiDAR (see \cref{tab:benchmark_leaderboard}) are presumed superior in this task, as explained in \cite{wu2022trajectory}. Although TCP reports a DS of 69 in \cite{wu2022trajectory}, reproductions vary notably: Jaeger \etal \cite{jaeger2023hidden} achieve 48 DS for their reproduced TCP, whereas our reproduction attains 58 DS. The CARLA leaderboard ensures repeatability via docker-based evaluations but does not guarantee reproducibility, as publicly released code and models can differ from the officially submitted models. These discrepancies align with experimental findings by Jaeger \etal (cf. Table 17 in \cite{jaeger2023hidden}), indicating that implementation details not explicitly highlighted in the original papers significantly influence leaderboard outcomes.
\begin{table*}[t]
 \caption{\textbf{Performance Comparison on the CARLA Public Leaderboard 1.0} \cite{carla_leaderboard} (accessed in Jun. 2025). We compare our approach (DTCP) with 6 top-ranked leaderboard submissions. Our model surpasses its reproduced baseline (TCP Reproduced) across all metrics and achieves the highest route completion among all competitors. Despite using only a monocular camera, without ensemble techniques or traffic rules sub-tasks, it has attained state-of-the-art performance while requiring the lowest execution time.}
  \label{tab:benchmark_leaderboard}
  \centering
  \resizebox{0.95\textwidth}{!}{
  \begin{tabular}{
    @{}
    >{\centering\arraybackslash}p{1cm}
    l
    >{\centering\arraybackslash}p{1.5cm}
    >{\centering\arraybackslash}p{1.5cm}
    >{\centering\arraybackslash}p{1.5cm}
    >{\centering\arraybackslash}p{2.5cm}
    >{\centering\arraybackslash}p{1.5cm}
    >{\centering\arraybackslash}p{1.5cm}
    >{\centering\arraybackslash}p{1.5cm}
    >{\centering\arraybackslash}p{1.5cm}
    @{}
    }
    \toprule
    Rank &
    Method & 
    \multicolumn{2}{c}{Sensor Inputs} & Ensemble & \multicolumn{1}{c}{\makecell{Auxiliary tasks \\traffic rules}} & \#Execution time &
    \multicolumn{1}{c}{\makecell{Driving\\score\\$\%,\uparrow$}} &  
    \multicolumn{1}{c}{\makecell{Route\\completion\\$\%,\uparrow$}} & 
    \multicolumn{1}{c}{\makecell{Infraction\\penalty\\$[0,1],\uparrow$}} \\
     & & \#Cameras & LiDAR & \\
    \midrule
    1 & ReasonNet \cite{shao2023reasonnet} & 5 & \cmark & \xmark & \cmark & $\sim117$ h & \bfseries 79.95 & 89.89 & \bfseries 0.89 \\
    2 & InterFuser \cite{shao2023safety} & 4 & \cmark & \xmark & \cmark & $\sim80$ h & 76.18 & 88.23 & 0.84 \\
    3 & TCP Ensemble \cite{wu2022trajectory} & 1 & \xmark & \cmark & \xmark & $\sim80$ h & 75.14 & 85.63 & 0.87 \\ 
    4 & DTCP (\textbf{ours}) & \tikzmarknode{startA}{1} & \xmark & \xmark & \xmark & \tikzmarknode{endB}{$\sim50$} h & 70.74  & \bfseries 95.60 & 0.75 \\ 
    5 & TF++ WP Ensemble \cite{jaeger2023hidden} & 1 &  \cmark & \cmark & \cmark & $\sim187$ h & 66.32 & 78.66 & 0.84 \\
    6 & LAV \cite{chen2022learning} & 4 & \cmark & \xmark & \cmark & $\sim394$ h & 61.85 & 94.46 & 0.64 \\
    - & TCP Reproduced (\textbf{our results}) & 1 & \xmark & \xmark & \xmark & $\sim50$ h & 58.56 & 83.14 & 0.69 \\
    \bottomrule
    \begin{tikzpicture}[overlay, remember picture]
      \draw[thick] 
        ([shift={(-8pt,2.5pt)}]startA.north) rectangle 
        ([shift={(24pt,-2pt)}]endB.south);
     \end{tikzpicture}
  \end{tabular}
  }
\end{table*}
\begin{table*}[t]
 \caption{\textbf{Infraction Analysis on CARLA Public Leaderboard 1.0} (accessed in Jun. 2025). We compare our approach (DTCP) against the top 6 baselines regarding infractions incurred. Our model records the second-lowest red light violation rate, achieving this without leveraging traffic-rule-specific auxiliary tasks. With minimal occurrences of agent blockage, off-road incidents, and deviations from the route, our agent demonstrates superior route completion rates, as reported in \cref{tab:benchmark_leaderboard}.}
 \label{tab:leaderboard_bench_infracs}
  \centering
  \resizebox{0.94\textwidth}{!}{
  \small 
  \begin{tabular}{
    @{}
    >{\centering\arraybackslash}p{1cm}
    l
    >{\centering\arraybackslash}p{1.5cm}
    >{\centering\arraybackslash}p{1.5cm}
    >{\centering\arraybackslash}p{1.5cm}
    >{\centering\arraybackslash}p{1.5cm}
    >{\centering\arraybackslash}p{1.5cm}
    >{\centering\arraybackslash}p{1.5cm}
    >{\centering\arraybackslash}p{1.5cm}
    >{\centering\arraybackslash}p{1.5cm}
    >{\centering\arraybackslash}p{1.5cm}
    @{}
  }
    \toprule
    Rank &
    Method &
    \multicolumn{1}{c}{\makecell{Collisions\\pedestrians\\$\text{\#km},\downarrow$}} &  \multicolumn{1}{c}{\makecell{Collisions\\vehicles\\$\text{\#km},\downarrow$}} & \multicolumn{1}{c}{\makecell{Collisions\\layout\\$\text{\#km},\downarrow$}} & \multicolumn{1}{c}{\makecell{Red light\\infractions\\$\text{\#km},\downarrow$}} & \multicolumn{1}{c}{\makecell{Stop sign \\infractions\\$\text{\#km},\downarrow$}} & \multicolumn{1}{c}{\makecell{Off-road \\infractions\\$\text{\#km},\downarrow$}} & \multicolumn{1}{c}{\makecell{Route \\deviations\\$\text{\#km},\downarrow$}} & \multicolumn{1}{c}{\makecell{Route \\timeouts\\$\text{\#km},\downarrow$}}  & \multicolumn{1}{c}{\makecell{Agent \\blocked\\$\text{\#km},\downarrow$}} \\
        \midrule
        \centering
        1 & ReasonNet \cite{shao2023reasonnet} & 0.02 & 0.13 & 0.01 & 0.08 & 0.00 & 0.04 & 0.00 & 0.01 & 0.33\\
        2 & InterFuser \cite{shao2023safety} & 0.04 & 0.37 & 0.14 & 0.22 & 0.00 & 0.13 & 0.00 & 0.01 & 0.43\\
        3 & TCP Ensemble \cite{wu2022trajectory} & 0.00 & 0.32 & 0.00 & 0.09 & 0.00 & 0.04 & 0.00 & 0.00 & 0.54 \\ 
        4 & DTCP (\textbf{ours}) & \tikzmarknode{startA}{0.00} & 0.14 & 0.06 & 0.05 & 0.17 & 0.04 & 0.00 & 0.08 & \tikzmarknode{endB}{0.10} \\ 
        5 & TF++ WP Ensemble \cite{jaeger2023hidden} & 0.00 & 0.50 & 0.00 & 0.01 & 0.00 & 0.12 & 0.00 & 0.00 & 0.71 \\ 
        6 & LAV \cite{chen2022learning} & 0.04 & 0.70 & 0.02 & 0.17 & 0.00 & 0.25 & 0.09 & 0.04 & 0.10 \\
        - & TCP Reproduced (\textbf{our results}) & 0.00 & 0.11 & 0.28 & 0.09 & 0.17 & 0.02 & 0.00 & 0.12 & 0.30\\ 
        \bottomrule
        \begin{tikzpicture}[overlay, remember picture]
          \draw[thick] 
            ([shift={(-14pt,2.5pt)}]startA.north) rectangle 
            ([shift={(16pt,-2pt)}]endB.south);
        \end{tikzpicture}
      \end{tabular}
      }
    \end{table*}
\newline \indent Despite not employing any ensemble model like TCP-Ens and TF++ WP or specific tasks for red light and stop sign detection like those found in ReasonNet and InterFuser, we recorded the second-lowest red light infraction rate as shown in \cref{tab:leaderboard_bench_infracs}. When compared to LAV \cite{chen2022learning}, our approach yields almost the same RC with a 0.11 higher IP, without using any multi-stage training with many pretrained modular components and the distillation framework utilized by LAV. Our agent experiences the lowest blocking rate without adopting any creeping strategy employed by TF++ WP \cite{jaeger2023hidden} to address the inertia problem observed in the IL agent \cite{codevilla2019exploring, chitta2022transfuser}, which tends to cause a stilled vehicle to remain inactive. Interestingly, in contrast to \cref{tab:lav_bench_infracs}, DTCP achieves a stop-sign infraction rate similar to the reproduced version while driving faster, as shown by $3\times$ fewer route timeouts and agent blocks. 
\newline \indent We also study the difference between the top six leaderboard submissions regarding the number of used sensors, model art, the inclusion of traffic rule auxiliary tasks, and the runtime duration. Both our model and TCP Ensemble employ only a front-view camera, whereas all other leading state-of-the-arts in \cref{tab:benchmark_leaderboard} use a LiDAR together with either multi-view cameras or a high-resolution monocular camera with a wide field of view (FOV), as in TF++ WP \cite{jaeger2023hidden}. In contrast to our model and TCP Ensemble \cite{wu2022trajectory}, both ReasonNet \cite{shao2023reasonnet} and Interfuser \cite{shao2023safety} utilize a traffic sign classifier to predict the traffic light state, determine if the ego agent is at an intersection, and identify an upcoming stop sign. TF ++ WP adopts many auxiliary tasks such as 2D depth, 2D and BEV semantic segmentation, and bounding box detection. For BEV semantic segmentation, TF++ WP predicts 11 classes, while for bounding box predictions, it distinguishes between 4 classes, including traffic light states and stop signs in both tasks. In addition to predicting a BEV semantic map, LAV \cite{chen2022learning} also leverages a brake predictor to handle traffic sign and hazardous stoppages using 3 camera-view images and an focusing-view (telephoto lens) image. Similar to LAV, ReasonNet and Interfuser also utilize a focus-view image as a strategy to capture distant traffic lights, as the small objects like traffic lights comprise only a few pixels in the whole input image, and could easily be overlooked in the prediction \cite{chen2024end}. Among all top models, our approach is the only one that does not rely on any ensemble or distillation framework, yet obtains competitive driving performance. Furthermore, our model requires $2.3\times$ less runtime compared to ReasonNet and thus achieves the lowest execution time by a significant margin from other competitors. Based on the results reported in \cite{wu2022trajectory}, we believe that using an ensemble model could potentially boost the driving performance of our model further.

\section{Conclusion}
In this research, we propose an interpretable end-to-end autonomous driving method by diversifying the feature representations. Our approach matches or outperforms state-of-the-art results on various benchmarks, particularly in reducing traffic infractions and achieving high route completion, while eschewing specialized auxiliary tasks. We demonstrate our model's enhanced adeptness at prioritizing critical safety considerations, thereby effectively reducing traffic violations and enhancing overall driving safety. Our findings illustrate the crucial role of feature diversity in bolstering the decision-making capabilities of autonomous driving agents, leading to a more streamlined and interpretable model. We show that improving interpretability is closely related to reducing infractions and, consequently, boosting driving performance. Notably, our proposed method has achieved state-of-the-art performance with minimized computational complexity on the CARLA public Leaderboard without the aid of ensemble techniques or extensive sensory inputs, which highlights its efficiency and scalability. As future work, we plan to integrate an ensemble model and evaluate its performance on the new CARLA Leaderboard.

{
    \small
    \bibliographystyle{ieeenat_fullname}
    \bibliography{main}
}

\clearpage
\setcounter{page}{1}
\maketitlesupplementary
\begin{center}
    {\large Mona Mirzaie \quad Bodo Rosenhahn}\\[0.5em]
    Institute for Information Processing\\
    Leibniz University Hannover, Germany\\[0.5em]
    {\tt\small mirzaie@tnt.uni-hannover.de}
\end{center}
\label{sec:appendix}
In this supplementary document, we first analyze failure cases in challenging scenarios and discuss the underlying causes (\cref{supSec:failure_case}). Next, we present a detailed analysis of our model’s behavior through visualizations, including activation maps that highlight key decision-making regions for critical traffic participants (\cref{supSec:visualzation_comparasion}). We then describe how interpretability metrics and mask generation are applied in our evaluations (\cref{supSec:gen_mask}), and subsequently provide a quantitative analysis of concept-level interpretability (\cref{supSec:c-iou}). Finally, we conduct an ablation study to assess the impact of hyperparameter choices on model performance  (\cref{supSec:w_divsity}).
\section{Failure Case Analysis} 
\begin{figure}[H]
  \centering
    \makebox[\linewidth][c]{%
    \hspace{2pt}
    \includegraphics[height=1.7cm]{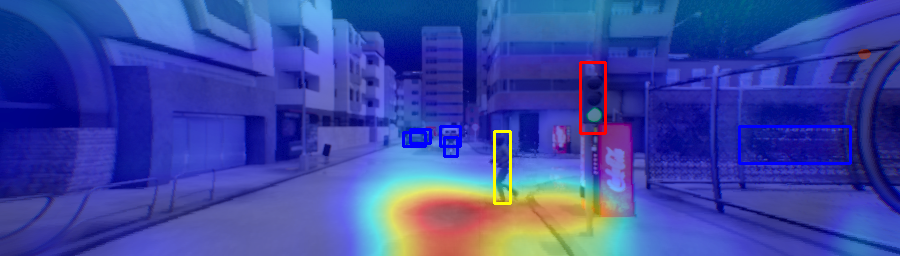}%
    \hspace{2pt}
    \includegraphics[height=1.7cm]{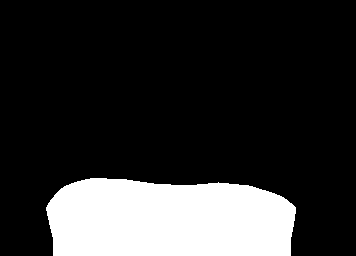}%
    \hspace{2pt}
    \includegraphics[height=1.7cm]{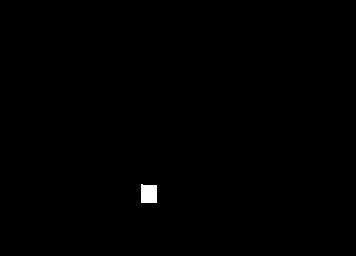}%
    \hspace{2pt}
    \includegraphics[height=1.7cm]{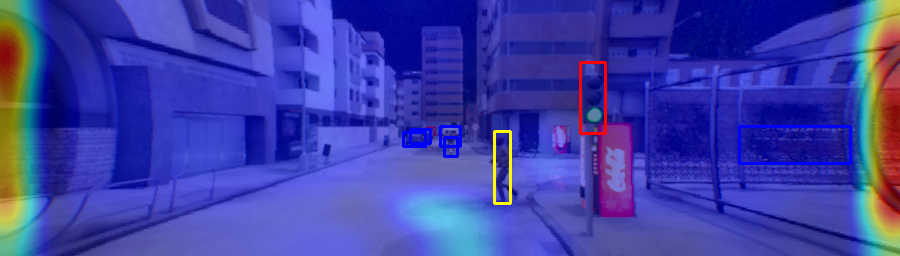}%
    \hspace{2pt}
   
}\vspace{2pt}

    \makebox[\linewidth][c]{%
    \hspace{2pt}
    \includegraphics[height=1.7cm]{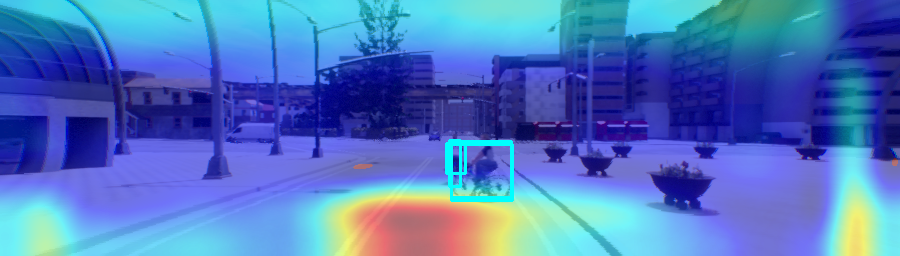}%
    \hspace{2pt}
    \includegraphics[height=1.7cm]{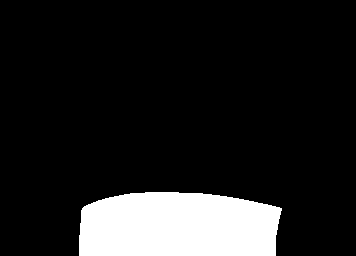}%
    \hspace{2pt}
    \includegraphics[height=1.7cm]{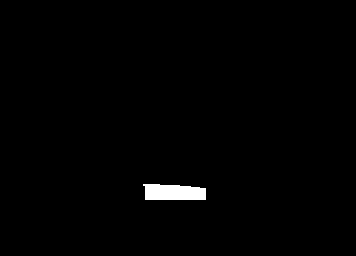}%
    \hspace{2pt}
    \includegraphics[height=1.7cm]{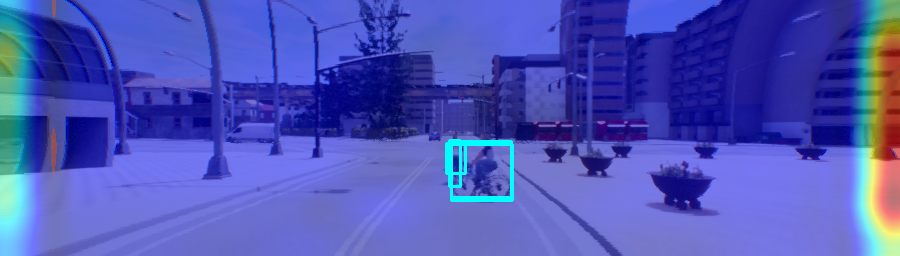}%
    \hspace{2pt}
}\vspace{2pt}

    \makebox[\linewidth][c]{%
    \hspace{2pt}
    \includegraphics[height=1.7cm]{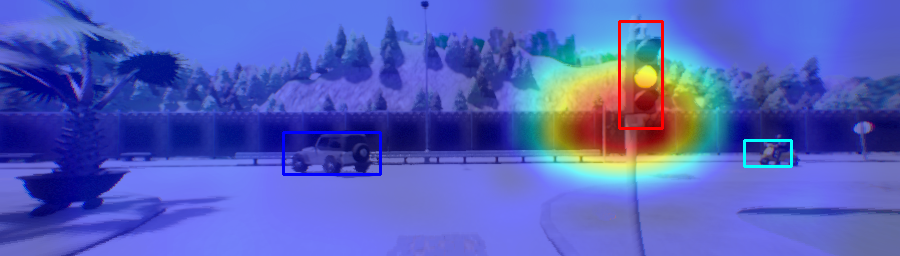}%
    \hspace{2pt}
    \includegraphics[height=1.7cm]{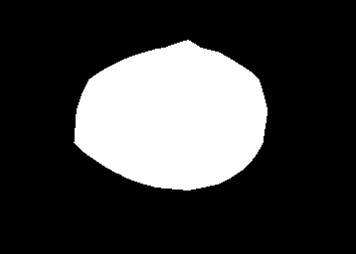}%
    \hspace{2pt}
    \includegraphics[height=1.7cm]{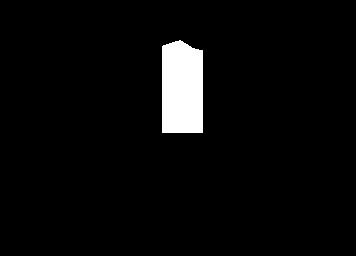}%
    \hspace{2pt}
    \includegraphics[height=1.7cm]{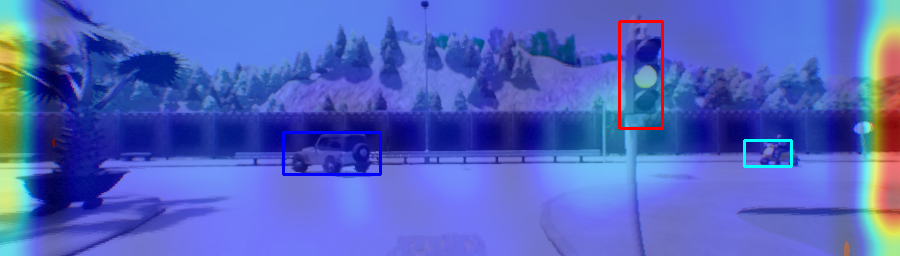}%
    \hspace{2pt}
}

  \caption{\textbf{Example of Failure Cases}. From left to right: heatmap of our model, heatmap binary mask, intersection of the bounding box and heatmap binary mask, and heatmap of the reproduced TCP.}
  \label{fig:failure_case}
\end{figure}

This section presents examples from \cref{fig:failure_case} demonstrating scenarios that lead to lower Intersection over Union (IoU) and Ground Truth Coverage (GTC) scores. Although our model successfully identifies relevant regions, the activation maps do not always fully overlap with ground-truth bounding boxes, resulting in lower scores. This partial coverage results in lower metric values, making it challenging to fully capture the interpretability advantages of our model based solely on these metrics.
\label{supSec:failure_case}

\section{Comparative Analysis of Activation Maps} 
\label{supSec:visualzation_comparasion}
\begin{figure}[H]
  \centering
  
    \makebox[\textwidth \vspace{0.3em}][c]{
   
  
     \hspace{1em} \textbf{DTCP (ours)} \hspace{17em}
   
   \textbf{TCP}
  }
  \makebox[\textwidth][c]{  
  
    \raisebox{3.3\height}{\textbf{(a)}}\hspace{1mm}
    \includegraphics[width=0.4\textwidth]{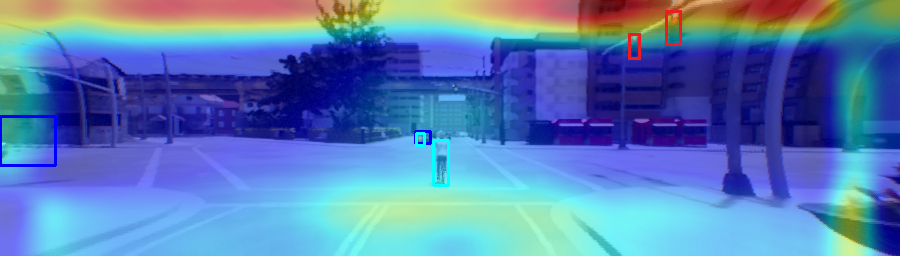}%
    \hspace{0.07mm}
   
    \includegraphics[width=0.4\textwidth]{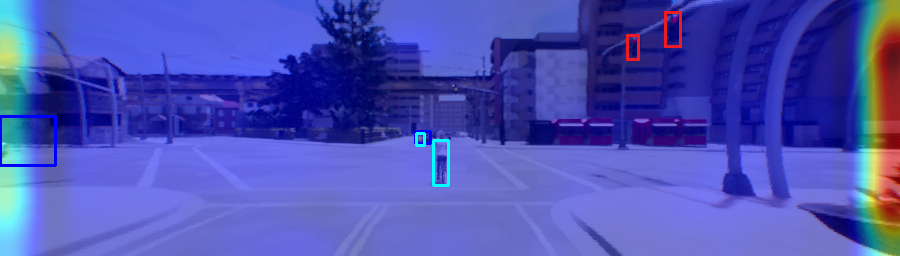}
  
  }
    \makebox[\textwidth][c]{  
  
     \raisebox{3.3\height}{\textbf{(b)}}\hspace{1mm}
    \includegraphics[width=0.4\textwidth]{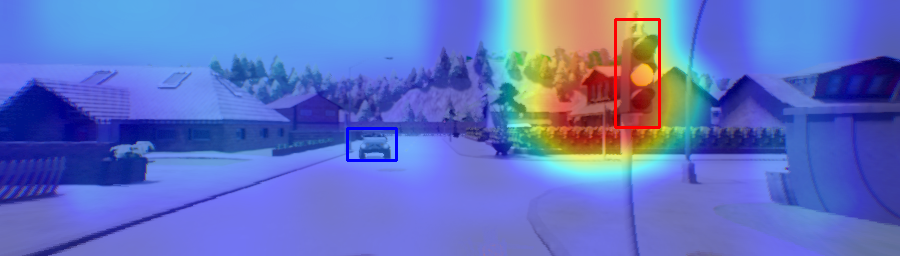}%
    \hspace{0.07mm}

    \includegraphics[width=0.4\textwidth]{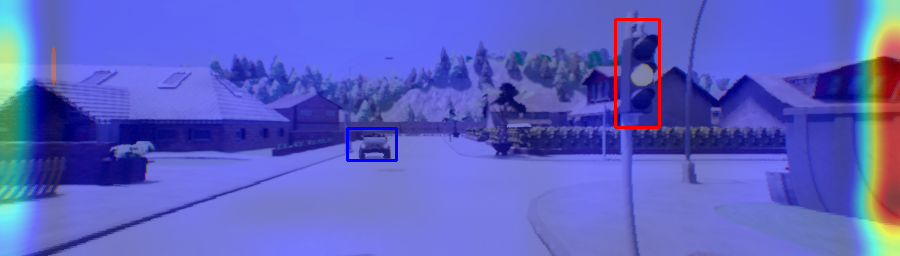}
  
  }
    \makebox[\textwidth][c]{  
  
     \raisebox{3.3\height}{\textbf{(c)}}\hspace{1mm}
    \includegraphics[width=0.4\textwidth]{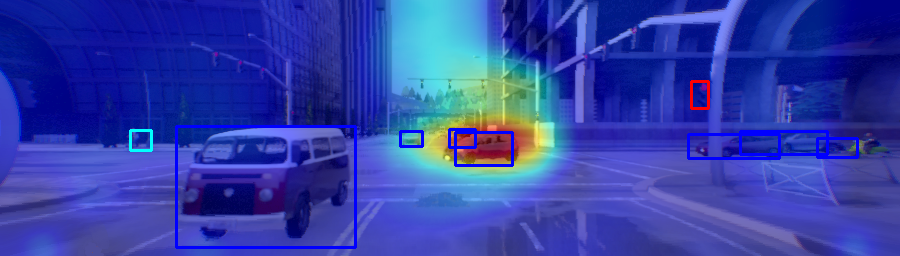}%
    \hspace{0.07mm}

    \includegraphics[width=0.4\textwidth]{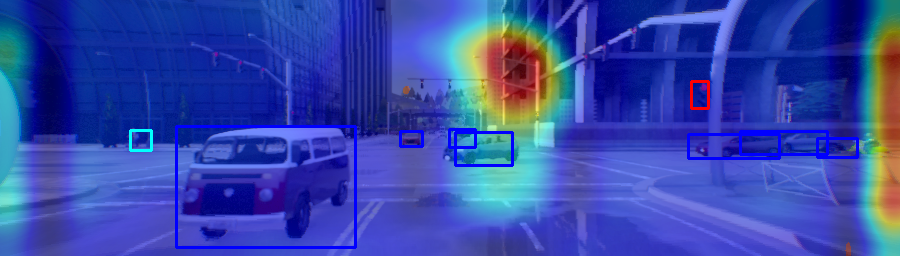}
  
  }

      \makebox[\textwidth][c]{  
  
     \raisebox{3.3\height}{\textbf{(d)}}\hspace{1mm}
    \includegraphics[width=0.4\textwidth]{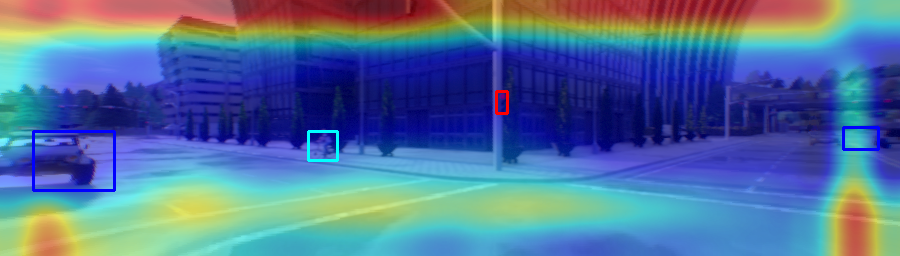}%
    \hspace{0.07mm}

    \includegraphics[width=0.4\textwidth]{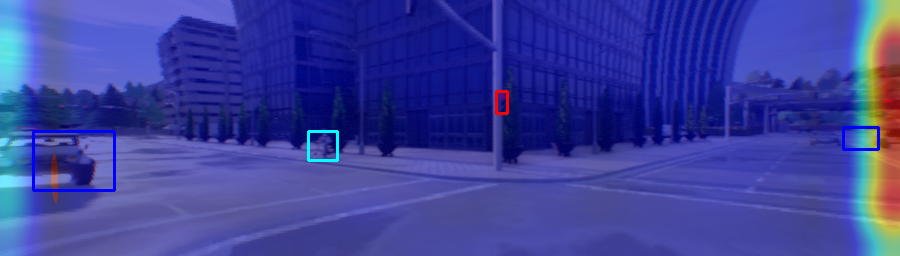}
  
  }

    \makebox[\textwidth][c]{  
  
     \raisebox{3.3\height}{\textbf{(e)}}\hspace{1mm}
    \includegraphics[width=0.4\textwidth]{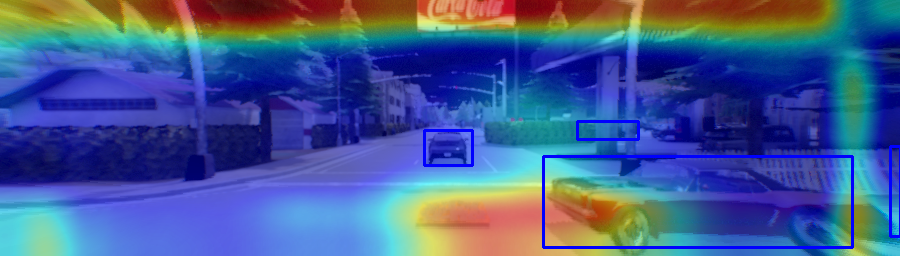}%
    \hspace{0.07mm}

    \includegraphics[width=0.4\textwidth]{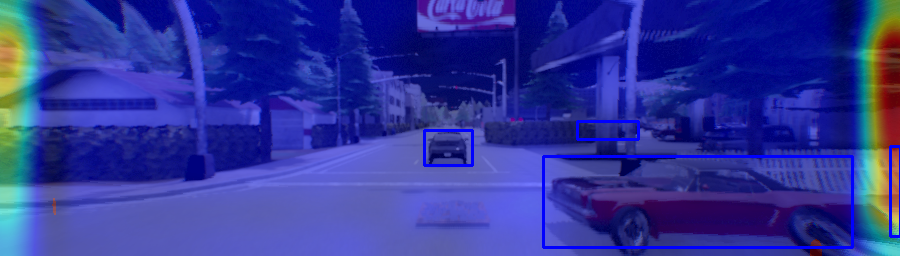}
  
  }

    \makebox[\textwidth][c]{  
  
     \raisebox{3.3\height}{\textbf{(f)}}\hspace{1mm}
    \includegraphics[width=0.4\textwidth]{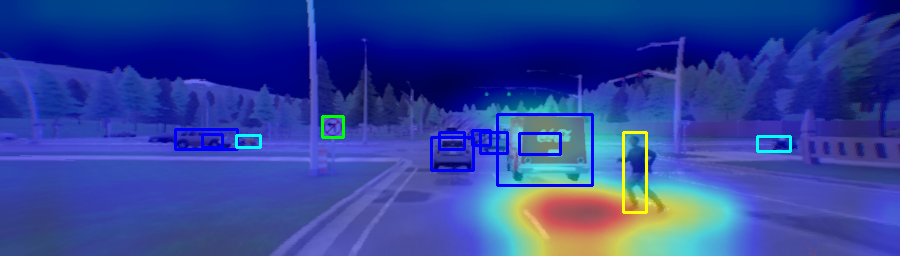}%
    \hspace{0.07mm}

    \includegraphics[width=0.4\textwidth]{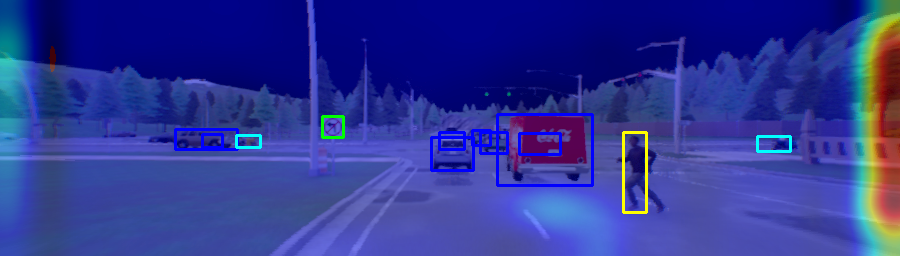}
  
  }
      \makebox[\textwidth][c]{  
  
     \raisebox{3.3\height}{\textbf{(g)}}\hspace{1mm}
    \includegraphics[width=0.4\textwidth]{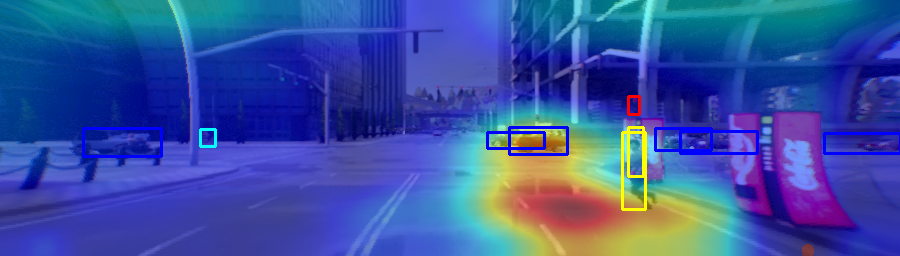}%
    \hspace{0.07mm}

    \includegraphics[width=0.4\textwidth]{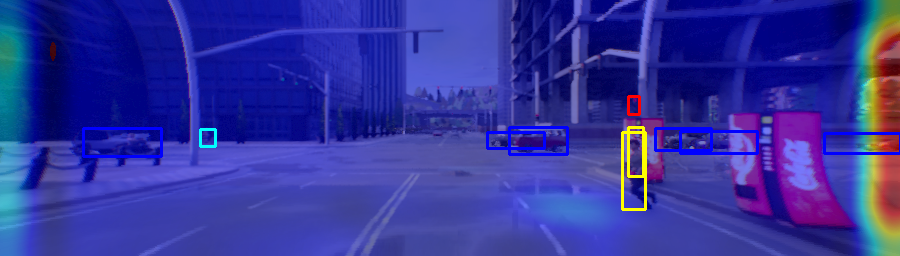}
  
  }

  \caption{\textbf{Comparison of activation maps}. Left: Heatmap of our model DTCP; right: heatmap of the reproduced TCP. Our model achieves superior performance by activating key regions relevant to driving decisions, thus enhancing interpretability.}
  \label{fig:examples_activation}
\end{figure}

In this section, we present additional qualitative analyses of activation maps generated using EigenCam \cite{muhammad2020eigen} for our proposed model DTCP compared to the baseline TCP \cite{wu2022trajectory}. As depicted in \cref{fig:examples_activation} \textbf{(a-g)}, DTCP effectively localizes attention on regions critical for driving decisions, significantly enhancing interpretability. Detecting traffic lights, especially those situated at higher and farther positions, is challenging due to their relatively small size in camera images. However, as illustrated in \cref{fig:examples_activation} \textbf{(a)}, DTCP effectively addresses this difficulty by accurately focusing its activation map within the bounding box of distant traffic lights. \cref{fig:examples_activation} \textbf{(b-c)} illustrates that our model assigns significant attention to yellow traffic lights. Specifically, in \textbf{(c)}, upon recognizing the yellow traffic light, the model effectively identifies vehicles within the intersection in our driving path as potential collision risks. As depicted in \cref{fig:examples_activation} \textbf{(d)}, our model carefully monitors the surrounding environment during turning maneuvers, proactively identifying potential risks from dynamic objects, such as cyclists (marked by a cyan bounding box) or pedestrians, that could unexpectedly enter our driving path. \cref{fig:examples_activation} \textbf{(e-g)} demonstrates that our model remains highly vigilant in detecting unexpected situations,  identifying hazards such as vehicles and pedestrians that violate traffic rules or run red lights, thereby reducing potential collision risks.

\section{Mask Generation and Interpretability Metrics} \label{supSec:gen_mask}
\begin{figure*}[htbp]
\centering
\begin{minipage}[b]{0.32\textwidth}
  \centering
  \includegraphics[width=\textwidth]{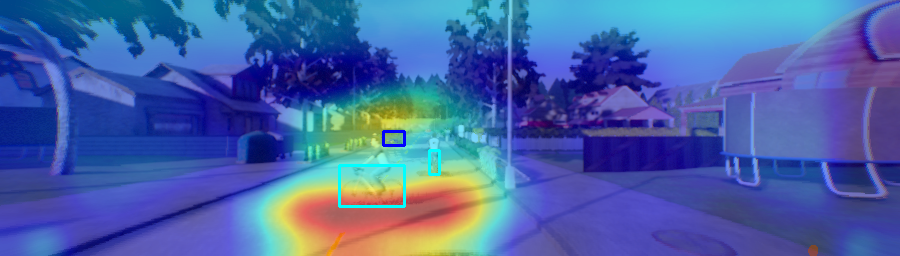}
\end{minipage}
\begin{minipage}[b]{0.32\textwidth}
  \centering
  \includegraphics[width=\textwidth]{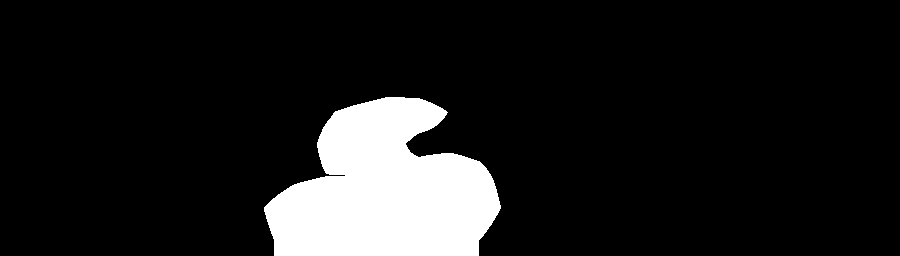}
\end{minipage}
\begin{minipage}[b]{0.32\textwidth}
  \centering
  \includegraphics[width=\textwidth]{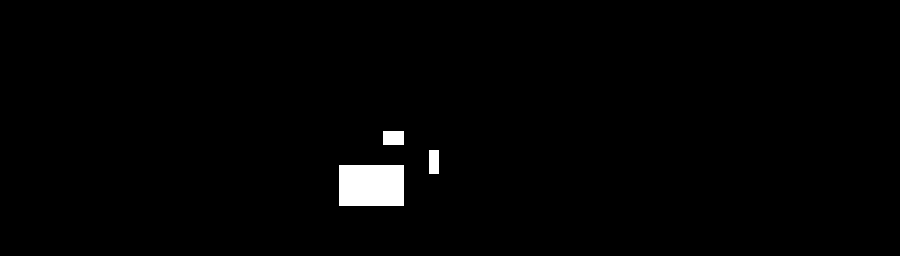}
\end{minipage}
\caption{\textbf{Example of Generated Binary Mask}. 
From left to right: heatmap with overlapping bounding boxes, 
binary mask derived from the heatmap, and binary mask of the bounding boxes.}
\label{fig:binary_mask}
\end{figure*}

For measuring interpretability using \emph{Shared Interest} metrics (IoU, GTC, SC) \cite{boggust2022shared}, we generate binary masks from object bounding boxes and our model’s heatmap, as illustrated in \cref{fig:binary_mask}. Ground-truth bounding boxes are created for categories most likely involved in infractions—pedestrians, cyclists, vehicles, and traffic lights.
\newline For a ground-truth set ($\mathcal{G}$) and a saliency set ($\mathcal{S}$), \emph{Shared Interest} metrics are defined as:
\begin{equation}
\mathrm{IoU}=\frac{|G \cap S|}{|G \cup S|}
\end{equation}
\begin{equation}
\mathrm{GTC}=\frac{|G \cap S|}{|G|}
\end{equation}
\begin{equation}
\mathrm{SC}=\frac{|G \cap S|}{|S|}
\end{equation}
\section{Concept-Level Interpretability} \label{supSec:c-iou}
To further demonstrate that DTCP attends more effectively to human-understandable concepts compared to the reproduced model, we compute driving-related semantic IoU, as reported in \cref{tab:c-iou}. Specifically, we measure IoU between model-generated saliency maps and key semantic classes: \textit{road}, \textit{roadline}, and \textit{sidewalk}. The advantage of DTCP in allocating greater attention to dynamic traffic participants and traffic lights is already illustrated in \cref{tab:heatmap_compare} of our main paper. Results in \cref{tab:c-iou} further confirm DTCP's superior ability to attend to driving-critical regions, reflecting human-like behavior essential for safe driving. 

\begin{table}[ht]                
  \caption{\textbf{Interpretability evaluation for driving-related concepts.} We report the average IoU scores. Our method allocates greater attention to decision-critical areas in driving scenarios.}
  \label{tab:c-iou}
  \centering
  \begin{tabularx}{\linewidth}{
    >{\centering\arraybackslash}p{2cm}    
    >{\centering\arraybackslash}p{3cm}    
    *{3}{>{\centering\arraybackslash}X}   
  }
    \toprule
    \multirow{2}{*}{\textbf{Routes}} &
    \multirow{2}{*}{\textbf{Method}} &
    \multicolumn{3}{c}{\textbf{IoU $\uparrow$}} \\ \cmidrule(lr){3-5}
    & & \textbf{Roadline} & \textbf{Road} & \textbf{Sidewalk} \\
    \midrule
    \multirow{2}{*}{Town01}
        & TCP \cite{wu2022trajectory} & 0.00 & 0.01 & 0.04 \\
        & DTCP (\textbf{ours})        & \textbf{0.01} & \textbf{0.30} & \textbf{0.21} \\ 
    \midrule
    \multirow{2}{*}{Town02}
        & TCP \cite{wu2022trajectory} & 0.00 & 0.02 & 0.03 \\
        & DTCP (\textbf{ours})        & \textbf{0.01} & \textbf{0.31} & \textbf{0.14} \\ 
    \midrule
    \multirow{2}{*}{Town03}
        & TCP \cite{wu2022trajectory} & 0.00 & 0.02 & 0.08 \\
        & DTCP (\textbf{ours})        & \textbf{0.02} & \textbf{0.41} &  \textbf{0.21} \\ 
    \midrule
    \multirow{2}{*}{Town05}
        & TCP \cite{wu2022trajectory} & 0.00 & 0.02 &  0.06 \\
        & DTCP (\textbf{ours})        & \textbf{0.02} & \textbf{0.40} & \textbf{0.11} \\ 
    \bottomrule
  \end{tabularx}
\end{table}

\section{Diversity Loss Weight Effect} \label{supSec:w_divsity}
\cref{tab:ablation_lambda_div} presents the impact of different loss weights ($\lambda_{\mathrm{div}}$) for our proposed diversity loss, as introduced in \cref{subsec:loss} of the main paper. We report the mean and standard deviation of the driving score over three evaluation runs in the CARLA simulator \cite{dosovitskiy2017carla} on the LAV benchmark \cite{chen2022learning}. As shown, setting $\lambda_{\mathrm{div}}=0.00005$ yields the best driving performance with the lowest standard deviation.
\begin{table}[!htbp]
  \caption{\textbf{Effects of Diversity Loss Weight}. An empirical analysis to determine the optimal value of $\lambda_{\mathrm{div}}$ in our model on the LAV Routes benchmark \cite{chen2022learning}.}
  \label{tab:ablation_lambda_div}
  \centering
  \resizebox{\textwidth}{!}{
  \small 
   \begin{tabular}{
    @{}
    >{\centering\arraybackslash}p{3cm}  
    >{\centering\arraybackslash}p{2cm}  
    >{\centering\arraybackslash}p{2cm}
    >{\centering\arraybackslash}p{2cm}
    >{\centering\arraybackslash}p{2cm}
    >{\centering\arraybackslash}p{2cm}
     >{\centering\arraybackslash}p{2cm}
    @{}
  }
    \toprule
    $\lambda_{\mathrm{div}}$ & $5 \times 10^{-1}$ & $5 \times 10^{-2}$ & $5 \times 10^{-3}$ & $5 \times 10^{-4}$ & $5 \times 10^{-5}$ & $5 \times 10^{-6}$ \\ 
    \midrule
    Driving Score & 42.87 $\pm$ 3.25 & 55.58 $\pm$ 6.44 & 48.22 $\pm$ 4.01 & 44.01 $\pm$ 1.29 & \textbf{60.42 $\pm$ 1.27} & 52.65 $\pm$ 1.78 \\ 
    \bottomrule
  \end{tabular}
  }
\end{table}

\end{document}